\documentclass[sigconf]{acmart}

\AtBeginDocument{%
  }

\copyrightyear{2026}
\acmYear{2026}
\setcopyright{cc}
\setcctype{by}
\acmConference[WWW '26] {Proceedings of the ACM Web Conference 2026}{April 13--17, 2026}{Dubai, United Arab Emirates.}
\acmBooktitle{Proceedings of the ACM Web Conference 2026 (WWW '26), April 13--17, 2026, Dubai, United Arab Emirates}
\acmISBN{979-8-4007-2307-0/2026/04}
\acmDOI{10.1145/3774904.3792259}

\usepackage{algorithm}
\usepackage{algorithmic}
\usepackage{multirow}
\usepackage{makecell}
\begin{document}

\title{CC-OR-Net: A Unified Framework for LTV Prediction through Structural Decoupling}
\author{Mingyu Zhao}
\authornote{Work performed during an internship at Meituan}
\email{mingyu087@ruc.edu.cn}
\affiliation{%
  \institution{Renmin University of China}
  \city{Beijing}
  \country{China}
}

\author{Haoran Bai}
\email{baihaoran04@meituan.com}
\affiliation{%
  \institution{Meituan}
  \city{Beijing}
  \country{China}
}

\author{Yu Tian}
\email{tianyu35@meituan.com}
\affiliation{%
  \institution{Meituan}
  \city{Beijing}
  \country{China}
}

\author{Bing Zhu}
\authornote{Corresponding author}
\email{zhubing04@meituan.com}
\affiliation{%
  \institution{Meituan}
  \city{Beijing}
  \country{China}
}

\author{Hengliang Luo}
\email{luohengliang@meituan.com}
\affiliation{%
  \institution{Meituan}
  \city{Beijing}
  \country{China}
}

\renewcommand{\shortauthors}{Mingyu Zhao, Haoran Bai, Yu Tian, Bing Zhu, and Hengliang Luo}
\begin{abstract}
Customer Lifetime Value (LTV) prediction, a central problem in modern marketing, is characterized by a unique zero-inflated and long-tail data distribution. This distribution presents two fundamental challenges: (1) the vast majority of low-to-medium value users numerically overwhelm the small but critically important segment of high-value "whale" users, and (2) significant value heterogeneity exists even within the low-to-medium value user base. Common approaches either rely on rigid statistical assumptions or attempt to decouple ranking and regression using ordered buckets; however, they often enforce ordinality through loss-based constraints rather than inherent architectural design, failing to balance global accuracy with high-value precision. To address this gap, we propose \textbf{C}onditional \textbf{C}ascaded \textbf{O}rdinal-\textbf{R}esidual Networks \textbf{(CC-OR-Net)}, a novel unified framework that achieves a more robust decoupling through \textbf{structural decomposition}, where ranking is architecturally guaranteed. CC-OR-Net integrates three specialized components: a \textit{structural ordinal decomposition module} for robust ranking, an \textit{intra-bucket residual module} for fine-grained regression, and a \textit{targeted high-value augmentation module} for precision on top-tier users. Evaluated on real-world datasets with over 300M users, CC-OR-Net achieves a superior trade-off across all key business metrics, outperforming state-of-the-art methods in creating a holistic and commercially valuable LTV prediction solution. 
\vspace{-8pt}
\end{abstract}


\vspace{-8pt}


\maketitle
\section{Introduction}
\begin{figure}[h!]
    \centering
    \includegraphics[width=0.7\columnwidth]{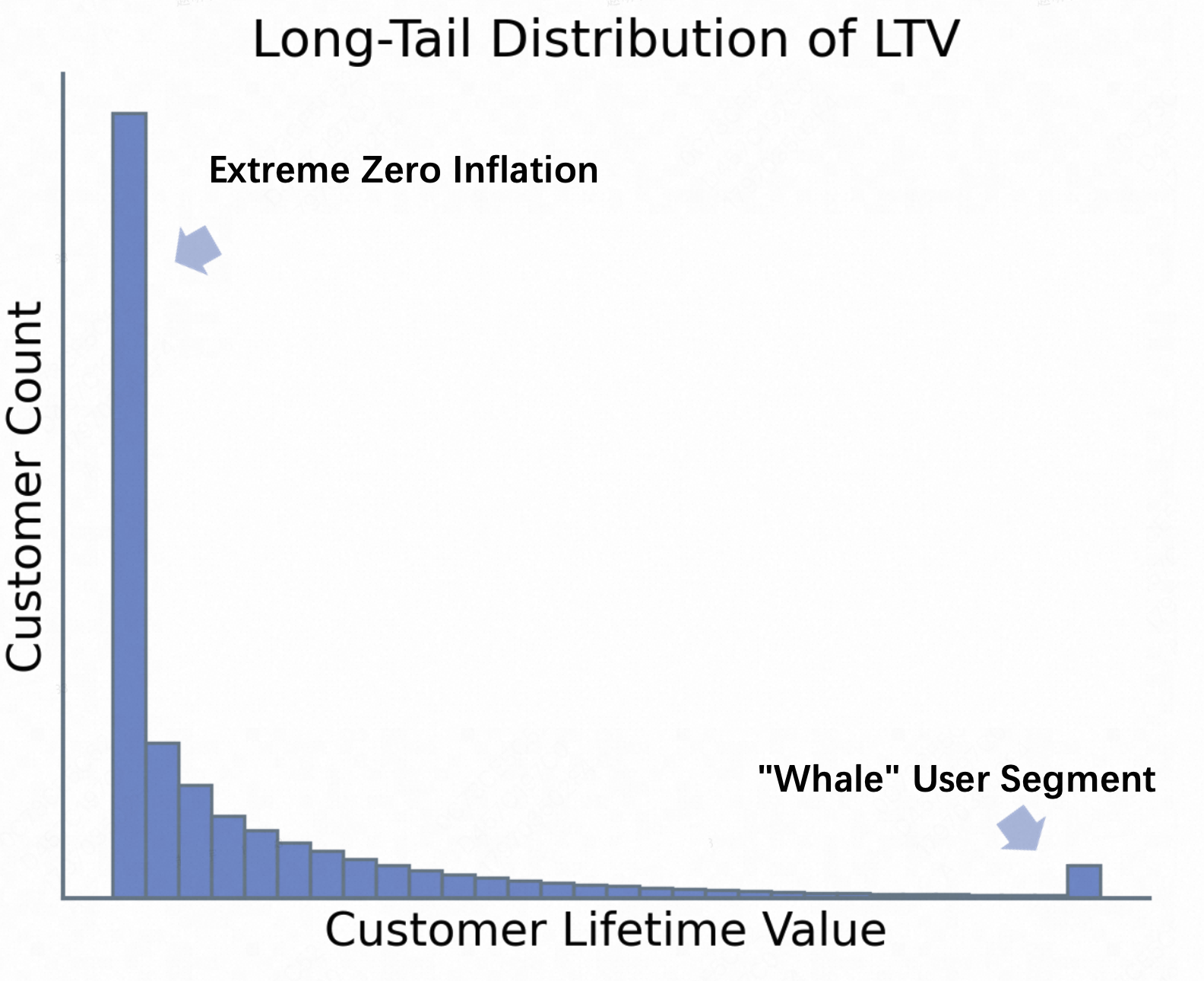}
    \caption{The challenging distribution of LTV, characterized by an Extreme Zero-Inflation peak, a severe Long-Tail, and a sparse but critical "Whale" User Segment.}
    \label{fig:ltv}
    \vspace{-5pt}
\end{figure}

In the modern Web ecosystem, which functions as a vast socio-economic platform, understanding and predicting user value is fundamental to sustainable platform growth and personalized user experiences. As a core challenge in \textbf{user modeling, personalization, and recommendation}, Customer Lifetime Value (LTV) prediction \citep{Gupta2006, Schmittlein1987} has become a cornerstone for Web platforms, directly influencing critical decisions in budget allocation, user segmentation, and ROI optimization. This paper addresses the scientific challenges of LTV prediction on the Web, where the core difficulty stems from its inherently challenging data distribution, as illustrated in Figure~\ref{fig:ltv}.

The typical LTV distribution is characterized by three distinct features that complicate modeling. First, an \textbf{Extreme Zero-Inflation} signifies a large population of non-paying or inactive users. Second, the non-zero values follow a severe \textbf{Long-Tail Distribution}, where the vast majority of users are concentrated in the low-value range, creating a significant imbalance that challenges models aiming for global regression accuracy. Third, the extreme end of this tail contains a sparse but business-critical \textbf{"Whale" User Segment}. Due to their rarity, their contribution to the overall loss function is minimal, causing most models to systematically under-predict their value—a critical failure in business applications on the Web.

Due to this complex data landscape, model performance cannot be judged by a single metric but must be assessed against a core \textbf{LTV prediction trilemma}, requiring models to simultaneously optimize three often-conflicting objectives: (1) \textbf{ranking quality}, to ensure users are correctly ordered by value; (2) \textbf{regression accuracy}, to predict the precise LTV for each user; and (3) \textbf{high-value precision}, to accurately identify and value the rare "whale" users. Existing approaches struggle to strike this balance. Methods that model the entire distribution with fixed statistical assumptions often fail to capture the nuances of the extreme tail. Conversely, while approaches using ordered bucket decomposition can improve ranking, they typically enforce ordinality through loss-based constraints rather than inherent architectural design. This indirect enforcement can be less robust and, when combined with modeling sub-distributions independently, leads to data fragmentation and inefficient learning, especially for the sparse high-value bucket. Furthermore, many such methods fail to explicitly isolate the zero-value population, a step that is critical for generating actionable marketing insights on Web platforms.

To address this gap, we introduce CC-OR-Net, a \textbf{novel unified framework} designed for large-scale Web platforms that structurally decouples ordinal ranking from regression. By transforming this "hard conflict" into a manageable "soft trade-off," our modular architecture systematically addresses the multi-objective problem. It synergistically combines specialized modules for ranking, regression, and high-value augmentation, guided by a principled loss function that balances these competing goals.In summary, our \textbf{Key Innovations} are:
\begin{enumerate}
    \item A \textbf{novel unified framework} that structurally decouples ordinal ranking from regression, featuring a scalable ordinal cascade with $O(K)$ complexity that is well-suited for industrial-scale deployment.
    \item A pioneering \textbf{attention-guided augmentation} strategy that precisely enhances learning for rare, high-value users, overcoming the limitations of generic re-sampling methods.
    \item The proposal of the \textbf{SVA metric}, a business-centric evaluation tool designed for user stratification tasks.
\end{enumerate}
\vspace{-7pt}
\section{Related Work}
The field of LTV prediction comprises a complex landscape of methods, yet none holistically address the intertwined challenges of ranking, regression, and high-value precision.

\textbf{Ordinal Regression and Long-tail Learning.} Preserving the rank-ordering of users by value is fundamental. While modern deep learning approaches like CORAL \citep{Cao2020} and CORN \citep{Shi2021Deep} achieve linear complexity, their monolithic designs are ill-suited to solving the LTV trilemma. Their unified structure prevents the integration of specialized modules—such as a dedicated enhancement mechanism for top-tier users—which is a core feature of our proposed architecture. Furthermore, standard long-tail learning techniques, such as re-sampling \citep{Chawla2002,He2009} or focal loss \citep{Lin2017}, are designed for nominal classification and often disrupt the critical ordinal structure inherent in value prediction.

\textbf{Specialized LTV Distributional Modeling.} Another line of work focuses on directly modeling the zero-inflated, long-tail distribution of LTV. Methods that assume specific statistical forms (e.g., ZILN \citep{Chamberlain2017}, Tweedie \citep{Jorgensen1987}) are often too rigid to capture real-world data complexities. Recent industrial-scale methods \citep{Li2022, Su2023, Yang2023, Zhang2023, Weng2024, Liu2024, HiLTV2025} have advanced scalability, but by focusing on fitting a specific distribution, they treat LTV prediction as a pure regression problem. This approach fundamentally ignores the inherent ranking information, which is critical for robust user stratification and is structurally guaranteed by our ordinal framework.

\textbf{Design Philosophy vs. Multi-Expert Systems.} Unlike multi-expert systems such as MMOE \citep{Ma2018} or ExpLTV \citep{Zhang2023} that use soft, learnable routing to assign samples to experts, CC-OR-Net employs a fixed, structurally-defined pipeline. We argue that this design is better suited for LTV prediction, as soft routing can become unstable when dealing with the sparse data characteristic of high-value users, leading to inconsistent expert assignments. In contrast, CC-OR-Net's fixed ordinal pipeline provides a robust, non-parametric routing mechanism that ensures stable stratification, allowing specialized modules to operate on a reliable foundation.

\textbf{Other Advanced Architectures.} Our survey also considered other advanced architectures, such as probabilistic models (NGBoost \citep{Duan2020NGBoost}, ZIGP \citep{Liu2018Zero}), hybrid models (DeepGBM \citep{Ke2019DeepGBM}), and Transformer models (MDLUR \citep{MDLUR2023}). However, their inherent design principles proved less suitable for addressing the LTV trilemma. For instance, Transformers require substantial feature engineering for heterogeneous data, and their attention mechanisms do not inherently align with the ordinal nature of the problem. Our preliminary investigations confirmed that without significant, problem-specific adaptations, these models did not offer a competitive performance trade-off.

To our knowledge, no prior work offers a similarly integrated solution validated at this scale. We select four representative categories of methods (traditional, ordinal, deep, and specialized distribution) for a comprehensive comparison. Other architectures showed inferior trade-offs in our experiments and are thus omitted from the main tables for brevity.
\vspace{-7pt}
\section{Methodology}
\subsection{Problem Formulation}

\begin{figure*}[t]
\centering
\includegraphics[width=\textwidth]{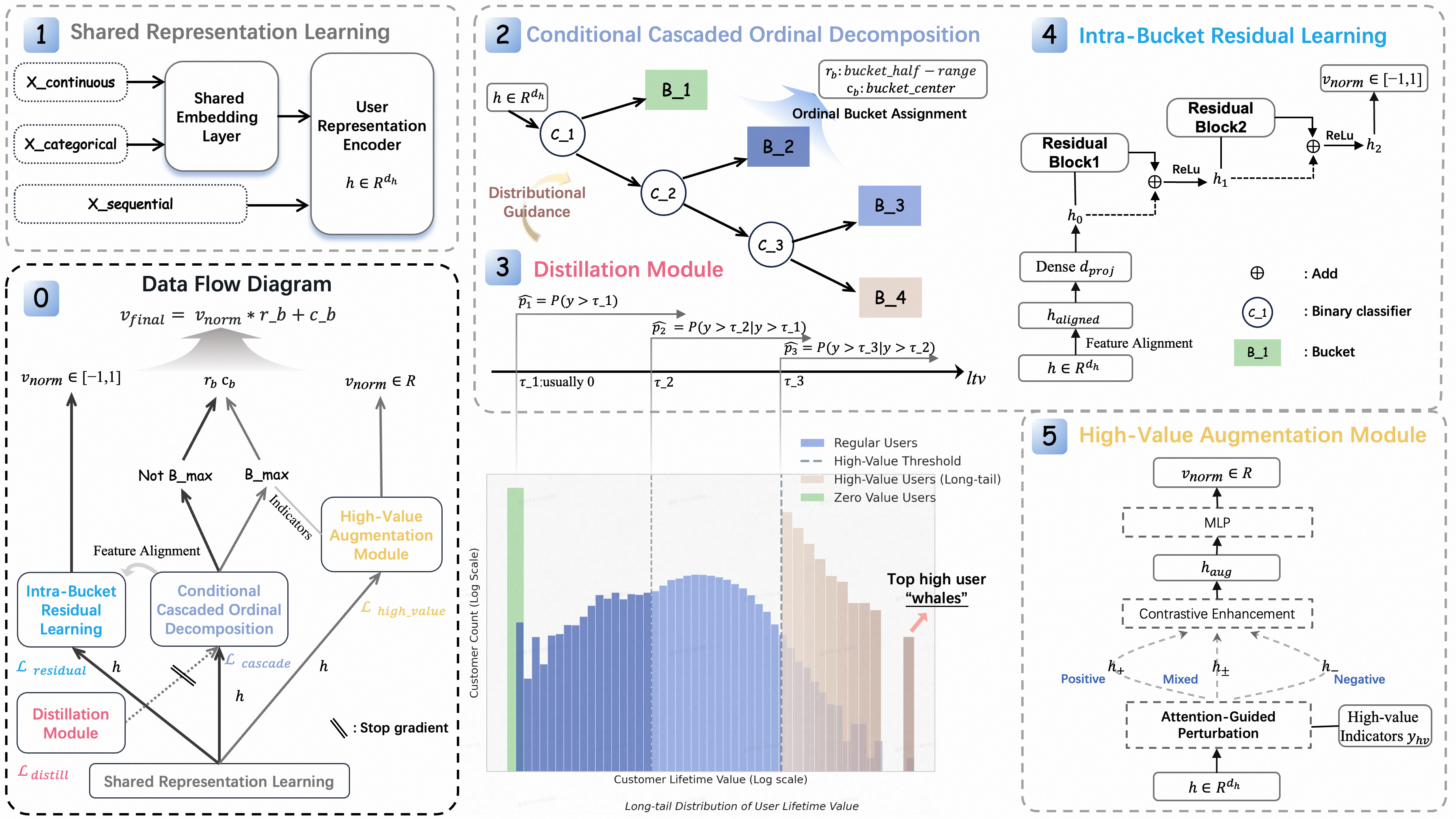}
\caption{CC-OR-Net Overall Architecture: Unified framework processing heterogeneous features through four specialized modules: (left) shared representation learning encoding multi-modal inputs, (center) conditional cascaded ordinal decomposition with fixed quantile boundaries, (right-top) intra-bucket residual learning for fine-grained regression, and (right-bottom) high-value augmentation module with attention-guided augmentation. The bottom panel displays the characteristic long-tail distribution with highlighted high-value segments, while the data flow diagram shows prediction integration across modules.}\label{fig:overall_architecture}
\end{figure*}

Given customer features $x \in \mathbb{R}^d$ and non-negative lifetime values $y \in \mathbb{R}^+$, LTV prediction faces three critical challenges: zero-inflated distributions with severe class imbalance, a natural ordinal structure requiring ordering preservation, and high-value sample scarcity where business-critical customers constitute sparse training data but drive majority revenue. While this work focuses on non-negative LTV, the framework is readily extensible to scenarios with negative values—which can arise from returns, service costs, platform subsidies, or marketing activities—by creating dedicated buckets for negative and zero users and adapting the regression head accordingly (e.g., by removing the final ReLU activation). We have validated this extension in our experiments, confirming its robustness. However, as this extension is beyond the core scope of this paper, we do not discuss it in further detail.

\textbf{Notation:} A comprehensive list of symbols is in Appendix Table \ref{tab:symbols}. Key symbols include the number of buckets $K$, bucket boundary thresholds $\tau_k$, and the conditional probability $p_k = P(\text{value} > \tau_k \mid \text{value} > \tau_{k-1}, x)$ predicted by the $k$-th classifier. By chaining these conditional exceedance probabilities, $\{p_k\}$, we obtain a valid K-way distribution that architecturally enforces ordinality. The probability of a user belonging to bucket $k$ is derived from the chain rule of probability, where a sample falls into bucket $k$ if and only if it passes the first $k-1$ classifiers but fails the $k$-th:
\begin{equation}
P(\text{bucket}=k | x) = 
\begin{cases} 
    1 - p_1 & k=1 \\
    (1 - p_k) \prod_{j=1}^{k-1} p_j & 1 < k < K \\
    \prod_{j=1}^{K-1} p_j & k=K
\end{cases}
\end{equation}
This formulation, in which $\prod_{j=1}^{k-1} p_j$ represents the probability of passing the first $k-1$ stages, inherently enforces an ordinal structure. Crucially, it guarantees a valid probability distribution whose components sum to 1, a property proven via a telescoping series in Appendix~\ref{sec:appendix_a}. This architectural guarantee is a key advantage over methods relying on learned constraints, as it is robust by design.
\vspace{-5pt}
\subsection{Overall Architecture}

Figure~\ref{fig:overall_architecture} presents the CC-OR-Net framework, which addresses the LTV trilemma through three specialized modules.

\textbf{Shared Representation Learning:} A multi-modal encoder processes heterogeneous features (categorical, continuous, sequential) into a unified representation $h \in \mathbb{R}^{d_h}$, which serves as input for all downstream modules.

\textbf{Conditional Cascaded Ordinal Decomposition:} This central component transforms the K-class ordinal problem into a sequence of $K-1$ binary classifications. Each classifier operates on samples that have survived the previous stages, with a Distillation Module providing distributional guidance. This design architecturally enforces ordering and achieves $O(K)$ complexity.

\textbf{Intra-Bucket Residual Learning:} This module captures fine-grained value variations within each bucket. Operating on context-aware features from the Feature Alignment mechanism, it outputs a normalized prediction $\hat{v}_{\text{norm}}$, which is then denormalized to obtain the final value $v_{\text{final}} = \hat{v}_{\text{norm}} \times r_b + c_b$.

\textbf{High-Value Augmentation Module:} This module focuses computational resources exclusively on the top user segment identified by the cascade. Using an attention-guided strategy, it generates augmented features ($h_{\text{aug}}$) to optimize prediction accuracy for "whales".

\textbf{Synergistic Integration:} The architecture implements a principled division of labor. We strategically employ \textbf{stop-gradient} operations between the cascade and distillation modules to prevent gradient conflicts and ensure stable learning, a known challenge in cascaded systems \citep{Frank2001}. This modular design enables scalable and effective LTV prediction.

\vspace{-5pt}

\subsection{Structural Ordinal Decomposition}
\label{sec:structural_ordinal_decomposition}
To ensure robust user ranking, often neglected by pure regression models on long-tail data, we reframe the K-class ordinal problem as a sequence of binary questions (e.g., \textit{'Is value > $\tau_1$?'}). Unlike independent cascaded classifiers \citep{Frank2001} prone to error propagation, CC-OR-Net is a unified, end-to-end architecture. Its key innovation is a probabilistic formulation (Eq. 1) that architecturally guarantees ordinal consistency (proven in Appendix~\ref{sec:appendix_a}).

Ordinal consistency implies a sample in bucket $k$ passed all prior $k-1$ thresholds. Eq. (1) enforces this by deriving bucket probabilities $P(\text{bucket}=k|x)$ from a sequence of conditional probabilities. This structural dependency ensures strict monotonicity by design, unlike learned constraints. It achieves $O(K)$ complexity and enables synergistic learning, facilitating industrial deployment.

\subsubsection{Cascaded Ordinal Learning: Implementation and Supervision}
The cascade employs $K-1$ binary classifiers for $K$ ordered buckets. We set $K=4$ to balance granularity with computational efficiency, though $K$ can be increased for finer-grained predictions. A key advantage of our design is that it converts the skewed regression task into a sequence of balanced binary classification sub-problems, ensuring that higher $K$ yields finer stratification without compromising performance or stability. We use \textbf{teacher-forcing} for training, where each classifier predicts marginal probabilities on the full dataset. During inference, samples traverse the cascade sequentially using conditional probabilities (Eq. 1).

\textbf{Bridging the Training-Inference Gap:} To mitigate training-inference discrepancies, we employ two mechanisms. First, a distillation loss ($L_{\text{distill}}$) regularizes the batch-level bucket distribution against the empirical one, significantly reducing distribution mismatch (e.g., 12.7\% reduction in Chi-squared on Domain 1). Second, fixed data-driven quantile thresholds ($\tau_k$) provide stable grounding. Our ablation study (Table \ref{tab:ablation}) confirms these components jointly improve accuracy and robustness.

\begin{algorithm}[h!]
\caption{Cascaded Structural Ordinal Decomposition}
\begin{algorithmic}[1]
\STATE \textbf{Shared representation learning:}
\STATE $h_{\text{shared}} = \text{Dense}(\text{ReLU}(\text{Dense}(h, d_{\text{hidden}})), d_{\text{output}})$

\STATE \textbf{Level-specific binary classifiers:}
\FOR{$i \in \{1, 2, \ldots, K-1\}$}
    \STATE $z_i = \text{ClassifierNetwork}(h_{\text{shared}}, \text{config}_i)$
    \STATE $p_i = \sigma(z_i)$ \COMMENT{Marginal prob. $P(\text{value} > \tau_i | x)$ for teacher-forcing}
\ENDFOR
\end{algorithmic}
\end{algorithm}
The `ClassifierNetwork` in Algorithm 1 is an MLP (details in Appendix). The cascade is trained with dual supervision:

\textbf{Point-wise Classification ($L_{\text{cascade}}$).} Each classifier $c_k$ is trained with a weighted BCE loss to predict if a user's value exceeds $\tau_k$. The target is $p_{\text{teacher},i,k} = \mathbb{I}(y_i > \tau_k)$, and various weights balance stages and class imbalance (see Appendix).
{
\begin{equation}
    L_{\text{cascade}} = \sum_{k=1}^{K-1} w_k \cdot \mathbb{E}_{i \in \mathcal{M}} \left[ \mathcal{L}_{\text{BCE}}(p_{\text{teacher},i,k}, \hat{p}_{i,k}) \right]
\end{equation}
}

\textbf{Distribution-level Distillation ($L_{\text{distill}}$).} A KL-divergence loss matches the predicted bucket distribution with the empirical one in a mini-batch to capture population structure. Its stability is controlled via temperature and loss weighting (see Appendix).
{
\begin{equation}
L_{\text{distill}} = \text{KL}(Q_{\text{true}} || Q_{\text{pred}})
\end{equation}
}
This dual supervision mechanism serves complementary objectives: $L_{\text{cascade}}$ ensures local accuracy at each specific decision boundary, while $L_{\text{distill}}$ provides global calibration to strictly align the predicted stratification with the true data distribution. We control the stability of distillation via temperature and loss weighting (see Appendix).

\vspace{-5pt}
\subsection{Feature Alignment Mechanism}
\label{sec:feature_alignment}
To effectively pass context from the ordinal cascade to the residual module, we designed a \textbf{Feature Alignment Mechanism}. This mechanism is motivated by the need to explicitly condition the fine-grained regression task on the coarse-grained ordinal output, thereby creating a context-aware input for the residual module. Instead of simply passing the shared features, this mechanism enriches them with both soft (probabilistic) and hard (categorical) context from the cascade and then uses a gated unit to dynamically filter and refine this information. 

\begin{algorithm}[h!]
\caption{Context-Aware Feature Alignment}
\label{alg:feature_alignment}
\begin{algorithmic}[1]
\STATE \textbf{Input:} Shared representation $h$, cascade probabilities $p_{\text{cascade}} = [p_1, \dots, p_{K-1}]$, predicted bucket index $\hat{b}$.
\STATE \textbf{Output:} Aligned features $h_{\text{aligned}}$.

\STATE \textbf{1. Feature Enrichment:}
\STATE $E(\hat{b}) \gets \text{EmbeddingLookup}(\text{bucket\_embeddings}, \hat{b})$ 
\STATE $h_{\text{base}} \gets \text{concat}(h, p_{\text{cascade}}, E(\hat{b}))$ 
\STATE \textbf{2. Gated Refinement (GLU-style):}
\STATE $g \gets \sigma(\text{Dense}_{\text{gate}}(h_{\text{base}}))$ 
\STATE $c \gets \text{ReLU}(\text{Dense}_{\text{content}}(h_{\text{base}}))$ 
\STATE $h_{\text{aligned}} \gets g \odot c$ 
\RETURN $h_{\text{aligned}}$
\end{algorithmic}
\end{algorithm}

The process, detailed in Algorithm~\ref{alg:feature_alignment}, creates a powerful, sample-specific feature representation for the subsequent regression task. While more complex mechanisms such as cross-attention were considered, we opted for a lightweight GLU-style gate as it offers an excellent trade-off between performance and computational cost—a critical factor for industrial-scale deployment.
\vspace{-5pt}

\subsection{Intra-Bucket Residual Learning}
While ordinal decomposition provides robust ranking, it inherently sacrifices granularity by collapsing continuous values. To bridge this gap and restore regression accuracy without compromising the ordinal structure, we introduce the Intra-Bucket Residual Learning module. Operating on aligned features $h_{\text{aligned}}$ (Section~\ref{sec:feature_alignment}), this module captures fine-grained value variations within each bucket. It employs a dual-block residual architecture \citep{He2016} with skip connections for precise, context-aware estimation while ensuring training stability.

\begin{algorithm}[h!]
\caption{Dual Residual Value Refinement Network}
\label{alg:residual_network}
\begin{algorithmic}[1]
\STATE \textbf{Input:} Aligned features $h_{\text{aligned}}$
\STATE \textbf{Output:} Normalized value prediction $v_{\text{norm}} \in [-1, 1]$
\STATE $h_0 \gets \text{Dense}(h_{\text{aligned}}, d_{\text{proj}})$
\STATE \textbf{Residual Block 1:}
\STATE $\text{res}_1 \gets \text{Dense}(\text{ReLU}(h_0), d_h)$
\STATE $h_1 \gets \text{ReLU}(\text{BN}(\text{res}_1) + h_0)$
\STATE \textbf{Residual Block 2:}
\STATE $\text{res}_2 \gets \text{Dense}(\text{ReLU}(h_1), d_r)$
\STATE $h_2 \gets \text{ReLU}(\text{res}_2 + \text{Dense}(h_1, d_r))$
\STATE $v_{\text{norm}} \gets \tanh(\text{Dense}(h_2, 1))$ 
\RETURN $v_{\text{norm}}$
\end{algorithmic}
\end{algorithm}

The module predicts a normalized value $\hat{v}_{\text{norm}} \in [-1, 1]$, which helps manage the scale of values within each bucket. To mitigate potential gradient vanishing issues caused by the predominance of zero-value samples in the first bucket, we implement dynamic label smoothing. Specifically, for samples in the first (zero-value) bucket, their normalized target label of 0 is replaced by a random value sampled uniformly from $[-\theta, \theta]$, where $\theta$ linearly decays from 1.0 to 0.1 during training, enabling gradual adaptation to the intra-bucket distribution. The final value prediction is obtained via bucket-specific denormalization:
{
\begin{equation}
v_{\text{final}} = \hat{v}_{\text{norm}} \times r_b + c_b
\end{equation}
}
where $r_b$ and $c_b$ are the bucket half-range and center, determined from training set quantiles. This bucket-specific denormalization, which scales the output based on the empirical distribution of the training data, provides a powerful adaptive capability. It ensures stable and effective residual learning across mini-batches, contributing significantly to the model's generalization performance.
\vspace{-5pt}
\subsection{Attention-Guided High-Value Augmentation}
To address the challenge of sparse "whale" users, for whom standard modules fall short and generic methods like re-sampling can disrupt the data distribution, we introduce a targeted augmentation module. It applies a novel attention-guided data augmentation strategy (Algorithm \ref{alg:enhancement}) exclusively to the user segment predicted to have the highest value ($\mathcal{H}$). This selective operation, based on the model's \textbf{predicted} bucket assignment, makes the module robust to minor misclassifications from the upstream cascade.

\begin{algorithm}[h!]
\caption{Attention-Guided Targeted Augmentation}
\label{alg:enhancement}
\begin{algorithmic}[1]
\STATE \textbf{Input:} Original features $h$ for a high-value sample
\STATE \textbf{Output:} Augmented features $h_{\text{aug}}$

\STATE \textbf{1. Generate Context-Aware Attention Weights:}
\STATE $h_{\text{stats}} \gets [\text{mean}(h), \text{std}(h), \max(h), \min(h)]$ 
\STATE $b_{\text{embed}} \gets E(\hat{b})$ 
\STATE $w \gets \text{sigmoid}(\text{MLP}([h; h_{\text{stats}}; b_{\text{embed}}]))$ 
\STATE \textbf{2. Apply Guided Perturbation:}
\STATE $\varepsilon \gets \mathcal{N}(0, \sigma_{\text{noise}}^2)$ 
\STATE $h_{\text{aug}} \gets h + \varepsilon \odot w$ 

\RETURN $h_{\text{aug}}$
\end{algorithmic}
\end{algorithm}

The augmented features $h_{\text{aug}}$ are processed by a dedicated dual-head network that predicts a regression value $\hat{v}_{\text{high}}$ and a confidence score $p_{\text{conf}}$ (Eq. 5). 
{
\begin{equation}
\hat{v}_{\text{high}} = \text{MLP}_{\text{reg}}(h_{\text{aug}}), \quad p_{\text{conf}} = \text{sigmoid}(\text{MLP}_{\text{conf}}(h_{\text{aug}}))
\end{equation}
}
This attention-guided augmentation acts as a form of \textbf{feature-level regularization}, similar to structured adversarial training. By selectively perturbing important features, it forces the model to learn more robust representations, significantly improving generalization for the sparse "whale" segment. Our preliminary experiments confirmed that this targeted approach, using a fixed noise level of $\sigma_{\text{noise}}=0.1$ (see Appendix \ref{sec:appendix_b}), is superior to simpler noise injection methods.
\vspace{-5pt}
\subsection{Loss Function Design}
\label{sec:loss_design}
The optimization is guided by a per-sample loss function $L_i$, which is then averaged over the batch. For any given sample $i$, the loss is a weighted sum of a main loss and a specialized high-value loss, controlled by an indicator function $\mathbb{I}(i \in \mathcal{H})$ that activates only for the highest-value bucket:
{
\begin{equation}
L_i = \gamma L_{\text{main}, i} + (1-\gamma) \mathbb{I}(i \in \mathcal{H}) L_{\text{high\_value}, i}
\end{equation}
}
where $\mathcal{H}$ is the set of samples in the highest value bucket. This formulation ensures the high-value loss is applied exclusively to the targeted segment. We set $\gamma=0.8$ to maintain strong global performance while still dedicating a significant learning signal to high-value precision. The main loss combines the cascade and residual objectives: $L_{\text{main}} = \alpha_1 L_{\text{cascade}} + \alpha_2 L_{\text{residual}} + \alpha_3 L_{\text{distill}}$. Here, $L_{\text{cascade}}$ and $L_{\text{distill}}$ provide dual supervision for the ranking task (Section~\ref{sec:structural_ordinal_decomposition}), while $L_{\text{residual}}$ targets general regression accuracy. To address the naturally left-skewed distributions within buckets (where lower values are more frequent), we employ a \textbf{value-weighted Mean Squared Error (MSE) loss} on the normalized predictions:
{
\begin{equation}
L_{\text{residual}} = \mathbb{E}[(\hat{v}_{\text{norm}} - v_{\text{norm}})^2 \times \text{sigmoid}(\beta \times v_{\text{norm}})]
\end{equation}
}
The sigmoid weighting term ($\beta=0.5$) acts as a soft, sample-specific weight. While other weighting functions (e.g., linear, exponential) were considered, the sigmoid form demonstrated the best empirical trade-off between stable gradient propagation and fitting effectiveness in our experiments. It dynamically increases the loss for higher-value samples within each bucket (where $v_{\text{norm}}$ is larger), forcing the model to pay more attention to the less frequent but more important tail-end of the intra-bucket distribution. This provides a simple yet effective re-weighting mechanism, analogous to how focal loss up-weights hard examples in classification.

For the critical high-value segment (bucket $\mathcal{H}$), we design a specialized loss, $L_{\text{high\_value}}$, with two components:

\textbf{1. Relative Regression Error ($L_{\text{relative}}$):} We use relative error, which aligns better with business-relevant percentage accuracy for high-value users.

\textbf{2. Confidence Loss ($L_{\text{conf}}$):} A Focal Loss-inspired term on the confidence head. We use Focal Loss with $\beta_{\text{focal}}=2$ because it allows the model to focus its learning on "hard" high-value samples where it is not yet confident, rather than easily achieving high confidence on all samples.
{
\begin{subequations}
\label{eq:loss_full_new}
\begin{align}
    L_{\text{high\_value}} &= \mathbb{E}_{i \in \mathcal{H}} \left[ L_{\text{conf},i} + \lambda_{\text{reg}} L_{\text{relative},i} \right] \\
    L_{\text{conf},i} &= -(1-p_{i}^{\text{conf}})^{\beta_{\text{focal}}} \log(p_{i}^{\text{conf}}) \\
    L_{\text{relative},i} &= \frac{|\hat{y}_i - y_i|}{\max(y_i, 1.0)}
    \end{align}
    \end{subequations}
}
\noindent where $\lambda_{\text{reg}}$ balances the two objectives. The denominator in $L_{\text{relative},i}$ is clipped at 1.0 to ensure numerical stability for low-value ground truth. This specialized loss is applied only to the outputs of the enhancement module, preventing gradient conflicts. Note that the confidence head $p_{\text{conf}}$ is a training-time mechanism to improve regression quality and is not used during inference to modify the final value prediction, ensuring a simple and direct output.
\vspace{-7pt}
\section{Experiments}
\subsection{Experimental Setup}
\subsubsection{Dataset Description}
Experiments are conducted on three large-scale industrial LTV datasets from different domains, each with tens to hundreds of millions of records exhibiting typical zero-inflation and long-tail properties where traditional regression struggles. To simulate production, datasets are split chronologically into training, validation, and test sets (40\% test).

\begin{table}[h]
\centering
\caption{Dataset Statistics}
\label{tab:dataset}
{
\setlength{\tabcolsep}{2pt}
\begin{tabular}{l|ccc}
\hline
\textbf{Characteristic} & \textbf{Domain 1} & \textbf{Domain 2} & \textbf{Domain 3} \\
\hline
Dataset Size & 248M records & 41M records & 33M records \\
Zero-Value Ratio & 33.6\% & 64.6\% & 45.8\% \\
\hline
\end{tabular}
}
\vspace{-5pt}
\end{table}

LTV values are partitioned into $K=4$ ordinal buckets. Bucket 1 isolates zero-value users (values in $[-10^{-6}, 10^{-6}]$). The remaining non-zero users are partitioned into three buckets using non-zero quantiles (e.g., 50th, 75th) to create balanced populations. This data-driven strategy ensures a consistent, reproducible setup for fair academic comparison. While production systems may use business-defined thresholds, our approach adapts to each dataset's unique distribution.

\subsubsection{Implementation Details}
Our implementation uses large-scale batching (100,000 samples) and parallel computation for efficient multi-core training. All experiments ran on 14-core CPUs. Detailed hyperparameters are in the Appendix.

\subsubsection{Baseline Methods}
We compare against four categories of methods: \textit{Traditional Methods} (XGBoost, two-stage approaches); \textit{Ordinal Regression} (CORAL \citep{Cao2020}, POCNN \citep{Niu2016}); \textit{Deep Learning} (DeepFM \citep{Guo2017}, MMOE-FocalLoss \citep{Ma2018,Lin2017}); and \textit{Specialized LTV} (ZILN \citep{Chamberlain2017}, MDME \citep{Li2022}, ExpLTV \citep{Zhang2023}, OptDist \citep{Weng2024}). All baselines use optimized hyperparameters with domain-specific adaptations. Detailed configurations are in the Appendix for reproducibility. Other architectures discussed in \textit{Related Work} were evaluated but omitted from tables due to non-competitive performance in our preliminary experiments.

\subsubsection{Evaluation Metrics}
We assess models across three dimensions: ranking, regression, and classification. To maintain clarity, we present the most critical metrics in the main text. Supplementary metrics for the cross-domain comparison, such as performance on non-zero subsets and detailed classification scores, are provided in the Appendix (Table \ref{tab:additional_metrics}).
\begin{itemize}
    \item \textbf{Ranking Quality:} We use the \textbf{GINI Coefficient} and \textbf{Spearman's Rank Correlation ($\rho$)} to measure the model's ability to correctly rank users. We also report these on the non-zero subset (GINI($y_i>0$), $\rho(y_i>0)$) to evaluate performance on the commercially active user base.
    \item \textbf{Regression Accuracy:} We evaluate point-wise prediction accuracy using \textbf{Normalized Mean Absolute Error (NMAE)}, \textbf{Mean Absolute Percentage Error (MAPE)}, \textbf{Average Mean Bias Error (AMBE)}, and \textbf{Normalized Root Mean Square Error (NRMSE)} to measure prediction bias.
    \item \textbf{Classification Performance:} We use standard metrics like \textbf{F1-score} (for zero/non-zero classification) and \textbf{Bucket Accuracy}. More importantly, we propose a business-centric metric, \textbf{Stratified Value Accuracy (SVA)}, to directly measure performance on user stratification tasks.
\end{itemize}

\subsubsection{Business-Centric Stratified Value Accuracy (SVA)}
Traditional metrics often rely on equal-frequency or equal-width bucketing, which are ill-suited for the extreme skew of LTV data. To address this, we propose \textbf{Stratified Value Accuracy (SVA)}, a business-centric metric that introduces a value-distribution-driven, adaptive threshold for stratification. Unlike conventional classification accuracy, SVA measures correct classification into three business-relevant strata: zero, low, and high-value, ensuring alignment with business value priorities. It is formally defined as:
{
\begin{equation}
\text{SVA} = \frac{1}{n}\sum_{i=1}^{n}\mathbb{I}(C(\hat{y}_i) = C(y_i))
\end{equation}
}
where $C(\cdot)$ maps a value $y$ to its stratum. While the threshold can be flexibly adjusted for specific business needs, in this paper, we use the median of positive values ($\tau_L = \text{median}(\{y_i \mid y_i > 0\})$) to ensure a fair, reproducible academic comparison. This adaptive approach makes SVA particularly robust for long-tailed value distributions.

\textbf{Implementation Note on Zero-Value Classification:} For fair comparison on classification metrics (SVA, F1), we apply a unified rule: for models with a direct non-zero probability (e.g., two-stage), we use that output; for pure regression models, a prediction is classified as non-zero if it exceeds a 0.1 threshold.

\begin{table*}[h]
\centering
\setlength{\tabcolsep}{1.2pt}
\small
\caption{Performance Comparison Across Three Domains \quad  (p$<$0.05)}
\label{tab:main_results}
\begin{tabular}{l|ccc|ccc|ccc|ccc|ccc|ccc|ccc}
\hline
\multirow{2}{*}{\textbf{Method}} & \multicolumn{3}{c|}{\textbf{GINI}} & \multicolumn{3}{c|}{\textbf{Spearman$\rho$}} & \multicolumn{3}{c|}{\textbf{AMBE}} & \multicolumn{3}{c|}{\textbf{NMAE}} & \multicolumn{3}{c|}{\textbf{MAPE}} & \multicolumn{3}{c|}{\textbf{NRMSE}} & \multicolumn{3}{c}{\textbf{SVA}} \\
\cline{2-22}
& D1 & D2 & D3 & D1 & D2 & D3 & D1 & D2 & D3 & D1 & D2 & D3 & D1 & D2 & D3 & D1 & D2 & D3 & D1 & D2 & D3 \\
\hline
\multicolumn{22}{l}{\textit{Traditional Methods}} \\
XGBoost & 0.709 & 0.243 & 0.397 & 0.730 & 0.207 & 0.416 & 14.85 & 8.38 & 80.41 & 0.823 & 1.106 & 1.122 & 9.67 & 3.26 & 31.59 & 2.174 & 5.957 & 6.849 & 35.77\% & 46.19\% & 29.96\% \\
Two-stage XGB & 0.774 & 0.423 & 0.482 & 0.683 & 0.184 & 0.372 & 10.32 & 4.38 & 71.10 & 0.847 & 1.444 & 1.465 & 18.60 & 9.56 & 34.78 & 2.342 & 6.373 & 4.337 & 48.19\% & 38.39\% & 39.75\% \\ 
\hline
\multicolumn{22}{l}{\textit{Ordinal Regression}} \\
CORAL & 0.750 & 0.370 & 0.471 & 0.742 & 0.232 & 0.423 & 14.18 & 5.81 & 63.84 & 0.846 & 1.688 & 1.138 & 14.51 & 6.23 & 64.27 & 2.386 & 6.925 & 4.041 & 52.34\% & 55.37\% & 44.15\% \\
POCNN & 0.751 & 0.447 & 0.479 & 0.731 & 0.210 & 0.433 & 21.13 & 6.40 & 74.50 & 0.849 & 1.771 & 1.185 & 21.41 & 6.82 & 74.89 & 2.137 & 5.321 & 3.604 & 47.06\% & 44.09\% & 43.46\% \\
\hline
\multicolumn{22}{l}{\textit{Deep Learning}} \\
DeepFM & 0.772 & 0.457 & 0.450 & 0.726 & 0.219 & 0.434 & 7.70 & 4.28 & 46.67 & 0.877 & 1.402 & 1.055 & 8.12 & 4.94 & 47.23 & 2.152 & 5.183 & 4.546 & 49.77\% & 50.72\% & 31.40\% \\
MMOE-FocalLoss & 0.788 & 0.487 & 0.461 & 0.732 & 0.245 & 0.433 & 8.73 & 3.36 & 46.97 & 0.790 & 1.334 & 1.039 & 9.17 & 4.04 & 47.55 & 2.293 & 5.299 & 3.757 & 39.83\% & 54.21\% & 39.45\% \\
\hline
\multicolumn{22}{l}{\textit{Specialized LTV}} \\
ZILN & 0.652 & 0.319 & 0.419 & 0.726 & 0.197 & 0.413 & 12.81 & 5.73 & 70.66 & 0.909 & 1.668 & 1.187 & 13.12 & 6.33 & 71.10 & 2.963 & 6.825 & 4.289 & 49.09\% & 32.16\% & 33.76\% \\
MDME & 0.794 & 0.476 & 0.488 & 0.749 & 0.223 & 0.438 & 5.95 & 1.25 & 27.19 & 0.785 & 1.029 & 1.090 & 6.71 & 1.34 & 27.71 & 2.393 & 5.132 & 4.118 & 60.55\% & 52.78\% & 51.41\% \\
ExpLTV & 0.781 & 0.468 & 0.485 & 0.755 & 0.261 & 0.434 & 5.12 & 1.15 & 24.35 & 0.798 & 0.991 & 1.012 & 6.35 & 1.37 & 43.88 & 2.255 & 4.989 & 4.050 & 57.15\% & 49.90\% & 38.10\% \\
OptDist & 0.785 & 0.465 & 0.480 & 0.737 & 0.212 & 0.420 & 14.05 & 2.33 & 39.06 & 0.815 & 1.292 & 1.014 & 14.55 & 3.13 & 39.74 & 2.148 & 5.519 & 3.987 & 55.42\% & 40.49\% & 32.87\% \\
\hline
\multicolumn{22}{l}{\textit{Our Method}} \\
CC-OR-Net & \textbf{0.803} & \textbf{0.490} & \textbf{0.501} & \textbf{0.761} & \textbf{0.278} & \textbf{0.442} & \textbf{4.85} & \textbf{0.90} & \textbf{21.18} & \textbf{0.776} & \textbf{0.986} & \textbf{0.982} & \textbf{5.53} & \textbf{1.25} & \textbf{21.06} & \textbf{2.093} & \textbf{4.834} & \textbf{3.316} & \textbf{67.01\%} & \textbf{60.50\%} & \textbf{53.89\%} \\
\hline
\end{tabular}
\end{table*}

\begin{table*}[t]
\centering 
\caption{Incremental Ablation Study Results for Domain 1\quad (p$<$0.05)}
\label{tab:ablation}
{\small
\setlength{\tabcolsep}{1.8pt} 
\begin{tabular}{l|ccccccccccc}
\hline
\textbf{Configuration} & \textbf{GINI} & \textbf{GINI($>$0)} & \textbf{Spearman$\rho$} & \textbf{Spearman$\rho$($>$0)} & \textbf{NMAE} & \textbf{MAPE} & \textbf{AMBE} & \textbf{NRMSE} & \textbf{SVA} & \textbf{F1} & \textbf{Bucket-Acc} \\
\hline
Baseline & 0.671 & 0.538 & 0.726 & 0.612 & 1.135 & 11.023 & 5.704 & 2.999 & 62.50\% & 0.695 & 0.486 \\
 +Cascaded Classification & 0.798 & 0.628 & 0.746 & 0.601 & 0.820 & 6.844 & 5.177 & 2.313 & 67.17\% & 0.786 & 0.556 \\
\hspace{0.5em} +Distillation Module & 0.794 & 0.615 & 0.759 & 0.619 & 0.767 & 5.830 & 6.257 & 2.144 & 66.55\% & 0.791 & 0.564 \\
\hspace{1.0em} +Residual Learning & \textbf{0.805} & \textbf{0.719} & 0.756 & 0.685 & \textbf{0.766} & 6.292 & 6.463 & 2.236 & 66.62\% & 0.802 & 0.553 \\
\hspace{1.5em} \makecell[l]{ +High-Value Augmentation \\ \qquad  (CC-OR-Net)} & 0.803 & 0.683 & \textbf{0.761} & \textbf{0.686} & 0.776 & \textbf{5.532} & \textbf{4.849} & \textbf{2.093} & \textbf{67.01\%} & \textbf{0.804} & \textbf{0.565} \\
\hline
\end{tabular}
}
\end{table*}

\subsection{Main Results}
\subsubsection{Overall Performance Comparison}
Our evaluation prioritizes the overall trade-off across ranking, regression, and business-centric metrics, rather than optimizing for a single metric. Table~\ref{tab:main_results} presents a comprehensive performance comparison across industrial-scale datasets, where CC-OR-Net consistently achieves a superior trade-off, with bold numbers indicating the best performance for each metric. Our method shows robust performance across datasets spanning tens to hundreds of millions of records, with all reported improvements being statistically significant (p$<$0.05).

While some specialized baselines excel on individual metrics, they often fail to address the trilemma holistically. For example, expert-based models like \textbf{MDME} and \textbf{ExpLTV} achieve strong regression accuracy, but their soft-routing mechanisms can struggle with the sparse high-value segment, resulting in less reliable stratification. Similarly, \textbf{OptDist}'s focus on distribution fitting sacrifices ordinal consistency, which our structural decomposition (Sec 3.3) guarantees. This design is critical for robust user stratification, as demonstrated by our superior SVA scores. Even strong traditional methods like \textbf{Two-stage XGB} exhibit high bias on the "whale" segment (high AMBE), showing that simply separating zero/non-zero prediction is insufficient.

Although GINI improvements may appear modest compared to some specialized baselines, for severely imbalanced LTV distributions, a high global GINI can mask deficiencies in critical high-value segments. Prior industrial-scale work has also noted that GINI may not reflect performance differences on imbalanced data \citep{Li2022}. CC-OR-Net's main strength is its ability to balance these trade-offs, maintaining high global ranking quality while achieving significant gains on business-centric metrics like SVA. This precise identification of impactful users is key to maximizing marketing ROI.
\vspace{-4pt}
\subsubsection{Incremental Architecture Components}
The ablation study in Table~\ref{tab:ablation} reveals the design philosophy behind CC-OR-Net by showing how each component systematically addresses the LTV trilemma.

\textbf{1. Ordinal Foundation and its Trade-offs:} The \textit{Cascaded Classification} module establishes a strong ordinal foundation, leading to a significant leap in ranking and stratification performance. Adding the \textit{Distillation Module} acts as a global regularizer, anchoring the model to the empirical data distribution. This improves overall regression stability but also reveals a key trade-off: enforcing global consistency can slightly dilute the model's focus on the tail-end of the ranking, a phenomenon further analyzed in Appendix~\ref{sec:appendix_distillation}.

\textbf{2. Exposing the Core Conflict and Targeted Correction:} The introduction of the \textit{Residual Learning} module exposes the central conflict of the LTV trilemma. While it pushes global ranking and regression accuracy to their peak, it simultaneously worsens the prediction bias for high-value users. This is not a flaw but an empirical confirmation of our core hypothesis: a globally-optimized regression objective is fundamentally ill-equipped to handle the precision requirements of the sparse "whale" segment.

\begin{table}[h!]
\centering
\caption{Impact of High-Value Augmentation on Top Bucket}
\label{tab:high_value_analysis}
\small
\begin{tabular}{l|ccc}
\hline
\textbf{Configuration} & \textbf{AMBE} &\makecell[l]{\textbf{NRMSE}  \\ \textbf{(Top Bucket)}} & \textbf{F1(Top Bucket)} \\
\hline
w/o Augmentation & 6.463 & 2.475 & 0.637 \\
CC-OR-Net (Full) & \textbf{4.849} & \textbf{2.132} & \textbf{0.674} \\
\hline
\textbf{Improvement} & \textbf{-25.0\%} & \textbf{-13.9\%} & \textbf{+5.8\%} \\
\hline
\end{tabular}
\end{table}
This necessitates a specialized correction mechanism, which is precisely the role of the final \textit{High-Value Augmentation} module. As shown in Table \ref{tab:high_value_analysis}, it acts as a targeted "corrector." By making a principled trade-off—accepting a negligible dip in global ranking—it delivers a dramatic \textbf{25.0\%} reduction in high-value bias (AMBE). This ability to sacrifice marginal global performance for a massive gain in business-critical precision is the core of our design philosophy and is a capability that strong baselines like Two-stage XGB lack (Table \ref{tab:main_results}). This targeted correction leads to the best overall balance, which is crucial for maximizing profitability in real-world marketing applications. The robustness of this design philosophy is further validated on other datasets (see Appendix Table~\ref{tab:domain3_ablation_complete}), demonstrating its ability to adapt to different data distributions.
\vspace{-4pt}
\subsubsection{Performance vs. Efficiency Trade-off}
\begin{figure}[h!]
    \centering
    \includegraphics[width=0.75\columnwidth]{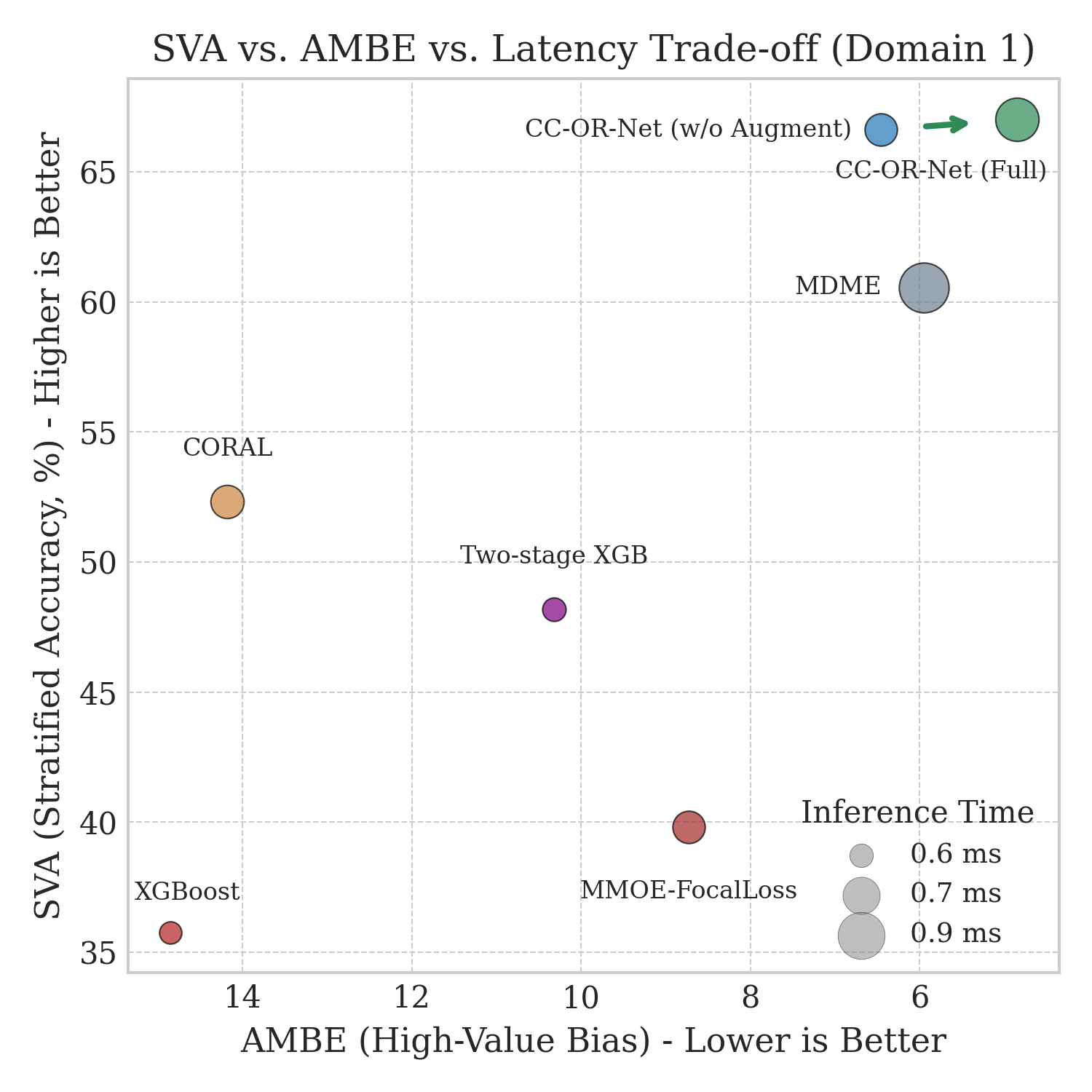}
  \caption{A 3D trade-off analysis on Domain 1. Axes represent overall accuracy (SVA) and high-value bias (AMBE), while bubble size indicates inference latency. The ideal region is top-right. CC-OR-Net demonstrates a superior balance, with the arrow showing the targeted improvement from our augmentation module.}
    \label{fig:tradeoff}
    \vspace{-5pt}
\end{figure}

Figure \ref{fig:tradeoff} visualizes the critical three-way trade-off between overall accuracy (SVA), high-value precision (AMBE), and inference latency. In this bubble chart, the ideal model occupies the top-right quadrant (high SVA, low AMBE) with the smallest possible bubble size (low latency). The chart clearly shows that while baselines are scattered, our CC-OR-Net variants occupy the most desirable positions. The arrow highlights the principled trade-off enabled by our High-Value Augmentation module: a minor increase in latency is strategically invested for a significant 25\% reduction in AMBE. This targeted improvement moves the model decisively into the optimal performance zone, achieving maximum business impact without sacrificing overall accuracy.

\vspace{-4pt}
\subsection{Business Impact Analysis}
To quantify direct business value, we use \textbf{Recall@k}, which measures the model's efficiency in identifying top-tier users ("whales") within a fixed marketing budget. It answers the question: "If we target the top k predicted users, what fraction of true whales do we capture?"

\begin{figure}[h!]
    \centering
    \includegraphics[width=0.75\columnwidth]{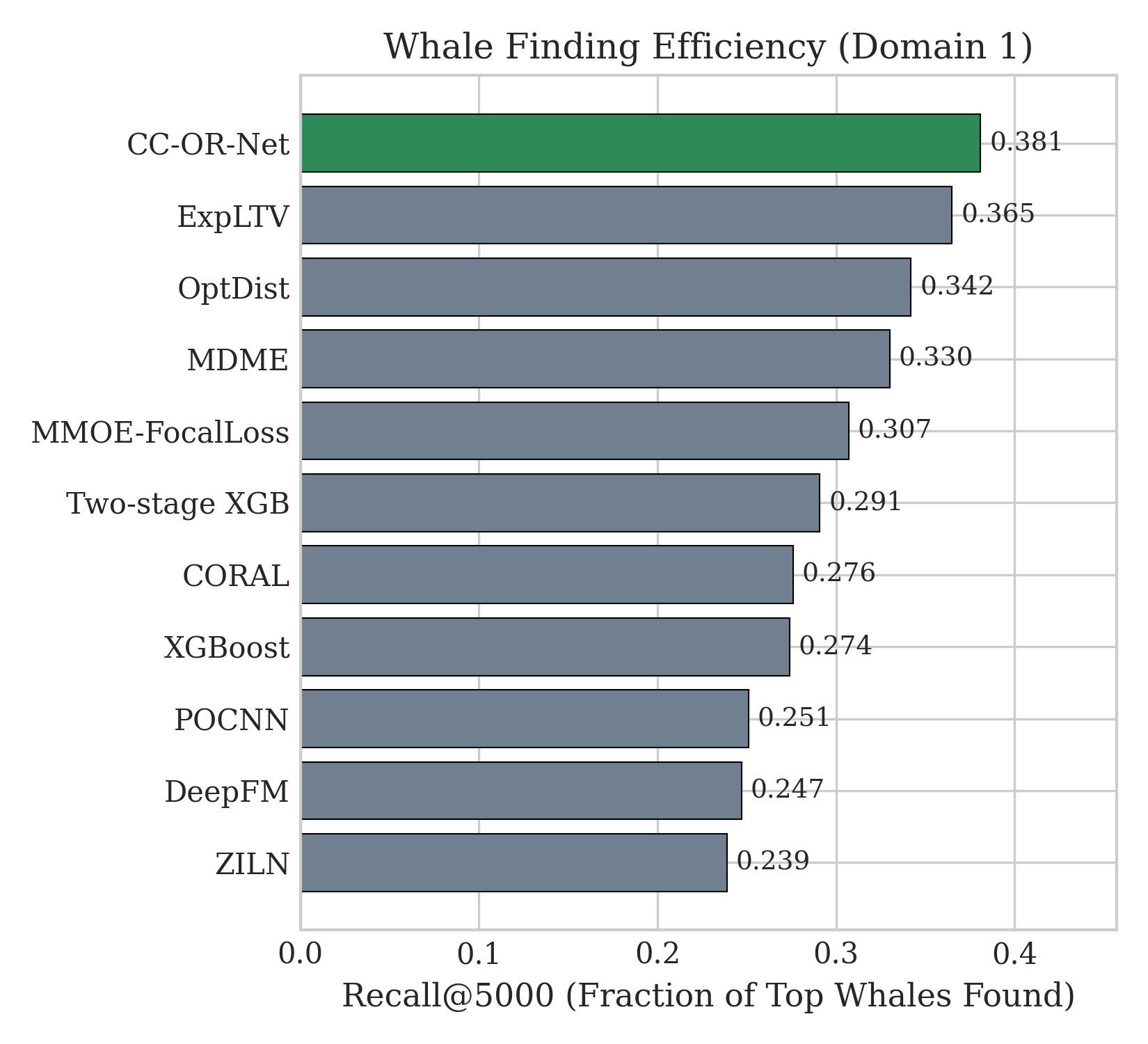}
    \caption{Whale finding efficiency on Domain 1, measured by Recall@5000. CC-OR-Net demonstrates a clear superiority in identifying the most valuable users within a fixed budget, a critical capability for maximizing marketing ROI.}
    \label{fig:whale_detection}
       \vspace{-5pt}
\end{figure}

Figure \ref{fig:whale_detection} compares Recall@5000 on Domain 1, simulating a campaign targeting 5,000 users. CC-OR-Net correctly identifies \textbf{38.1\%} of true "whales" in this cohort, a significant lead over strong baselines like ExpLTV (36.5\%) and OptDist (34.2\%). This superior "whale" detection capability directly translates to higher campaign ROI.

\vspace{-4pt}
\subsection{Computational Efficiency Analysis}

\begin{table}[h!]
\centering
\caption{Computational Efficiency Comparison}
\label{tab:efficiency}
{\small
\setlength{\tabcolsep}{2pt}
\begin{tabular}{l|cccc}
\hline
\textbf{Method} & \textbf{Inference} & \textbf{Peak} & \textbf{Parameters} & \textbf{FLOPs} \\
 & \textbf{Time} &\textbf{Memory*} & & \\
\hline
XGBoost & 0.54 ms & 67.38 GB & - & - \\
Two-Stage & 0.55 ms & 69.76 GB & 1.44 M & 2.95 M \\
CORAL & 0.65 ms & 67.38 GB & 1.10 M & 2.27 M \\
MMOE-FocalLoss & 0.64 ms & 69.10 GB & 0.90 M & 1.87 M \\
MDME & 0.90 ms & 75.13 GB & 1.24 M & 2.55 M \\
\hline
CC-OR (w/o Augmentation) & 0.64 ms & 72.62 GB & 0.93 M & 1.89 M \\
CC-OR (Full) & 0.79 ms & 78.84 GB & 1.04 M & 2.06 M \\
\hline
\multicolumn{5}{l}{\footnotesize{*Peak process memory for a batch of 100,000 samples, including data.}}
\end{tabular}
}
\end{table}
Table~\ref{tab:efficiency} shows the practical scalability of our approach. CC-OR-Net (w/o augmentation) is highly efficient, with latency matching fast baselines like MMOE (0.64ms). The reported \textbf{Peak Memory} is for a large batch (100,000 samples), which aids in deployment planning. The full CC-OR-Net incurs only a modest increase in latency (0.79ms) and memory, with the extra 0.15ms and ~6GB used for the High-Value Augmentation Module. As shown in our ablation study (Table \ref{tab:high_value_analysis}), this targeted investment yields a substantial \textbf{25.0\%} reduction in AMBE, a principled trade-off that allocates resources for maximum business impact. The architecture's parallelizable and modular design ensures high throughput and maintainability, supporting robust industrial deployment.
   \vspace{-7pt}
\section{Conclusion}
This paper introduced CC-OR-Net, a unified framework that resolves the LTV prediction trilemma of balancing ranking, regression, and high-value precision. By structurally decoupling these competing objectives via its modular architecture, CC-OR-Net achieves a superior performance trade-off on massive industrial datasets. Its principled design significantly reduces high-value prediction bias and improves stratification accuracy, aligning with key business goals. Multiple variants of the proposed model have been successfully deployed and integrated into production on Meituan's platforms, further validating its broad commercial utility. Beyond LTV, the CC-OR-Net paradigm of structural decomposition with targeted augmentation is applicable to other regression problems characterized by long-tail distributions, ordinal structure, and critical sparse samples, such as predicting user engagement or financial risk. It offers a principled solution for domains requiring a balance between global performance and precision on a vital minority.


\bibliographystyle{ACM-Reference-Format}
\bibliography{sample-base}


\appendix

\begin{table*}[t]
\centering
\setlength{\tabcolsep}{2.5pt}
\small
\caption{Additional Performance Metrics Across Three Domains\quad  (p$<$0.05)}
\label{tab:additional_metrics}
\begin{tabular}{l|ccc|ccc|ccc|ccc}
\hline
\multirow{2}{*}{\textbf{Method}} & \multicolumn{3}{c|}{\textbf{Bucket-Acc}} & \multicolumn{3}{c|}{\textbf{GINI($>$0)}} & \multicolumn{3}{c|}{\textbf{Spearman $\rho$ ($>$0)}} & \multicolumn{3}{c}{\textbf{F1}} \\
\cline{2-13}
& D1 & D2 & D3 & D1 & D2 & D3 & D1 & D2 & D3 & D1 & D2 & D3 \\
\hline
\multicolumn{13}{l}{\textit{Traditional Methods}} \\
XGBoost & 0.395 & 0.223 & 0.244 & 0.611 & 0.134 & 0.308 & 0.589 & 0.069 & 0.300 & 0.612 & 0.342 & 0.531 \\
Two-stage XGB & 0.401 & 0.331 & 0.295 & 0.645 & 0.485 & 0.397 & 0.655 & 0.385 & 0.395 & 0.705 & 0.462 & 0.636 \\
\hline
\multicolumn{13}{l}{\textit{Ordinal Regression}} \\
CORAL & 0.486 & 0.383 & 0.398 & 0.660 & 0.297 & 0.399 & 0.661 & 0.316 & 0.407 & 0.785 & 0.446 & 0.653 \\
POCNN & 0.409 & 0.377 & 0.341 & 0.663 & 0.440 & 0.347 & 0.654 & 0.310 & 0.392 & 0.640 & 0.438 & 0.598 \\
\hline
\multicolumn{13}{l}{\textit{Deep Learning}} \\
DeepFM & 0.450 & 0.272 & 0.262 & 0.654 & 0.476 & 0.330 & 0.639 & 0.307 & 0.389 & 0.614 & 0.436 & 0.540 \\
MMOE-FocalLoss & 0.370 & 0.302 & 0.313 & 0.671 & 0.481 & 0.327 & 0.651 & 0.389 & 0.391 & 0.550 & 0.425 & 0.571 \\
\hline
\multicolumn{13}{l}{\textit{Specialized LTV}} \\
ZILN & 0.409 & 0.235 & 0.187 & 0.566 & 0.144 & 0.224 & 0.651 & 0.326 & 0.374 & 0.659 & 0.363 & 0.463 \\
MDME & 0.520 & 0.372 & 0.446 & 0.674 & 0.529 & 0.333 & 0.627 & 0.352 & 0.251 & 0.789 & 0.444 & 0.643 \\
ExpLTV & 0.475 & 0.342 & 0.408 & 0.658 & 0.465 & 0.321 & 0.647 & 0.342 & 0.395 & 0.719 & 0.438 & 0.626 \\
OptDist & 0.422 & 0.261 & 0.252 & 0.664 & 0.519 & 0.358 & 0.657 & 0.391 & 0.403 & 0.741 & 0.387 & 0.538 \\
\hline
\multicolumn{13}{l}{\textit{Our Method}} \\
CC-OR-Net & \textbf{0.565} & \textbf{0.406} & \textbf{0.468} & \textbf{0.683} & \textbf{0.542} & \textbf{0.412} & \textbf{0.686} & \textbf{0.400} & \textbf{0.417} & \textbf{0.804} & \textbf{0.509} & \textbf{0.680} \\
\hline
\end{tabular}
\end{table*}

\begin{table*}[h!]
\centering
\setlength{\tabcolsep}{1.8pt}
\small
\caption{Domain 3 Incremental Ablation Study Results \quad  (p$<$0.05)}
\label{tab:domain3_ablation_complete}
\begin{tabular}{l|ccccccccccc}
\hline
\textbf{Configuration} & \textbf{GINI} & \textbf{GINI($>$0)} & \textbf{Spearman$\rho$} & \textbf{Spearman$\rho$($>$0)} & \textbf{NMAE} & \textbf{MAPE} & \textbf{AMBE} & \textbf{NRMSE} & \textbf{SVA} & \textbf{F1} & \textbf{Bucket-Acc} \\
\hline
Baseline & 0.419 & 0.282 & 0.389 & 0.312 & 1.292 & 41.906 & 41.036 & 6.309 & 40.88\% & 0.586 & 0.394 \\
 +Cascaded Classification & 0.481 & 0.363 & 0.426 & 0.378 & 1.070 & 25.304 & 22.493 & 4.255 & 53.79\% & 0.672 & 0.454 \\
\hspace{0.5em} +Distillation Module  & 0.494 & 0.367 & 0.431 & 0.371 & 1.039 & 23.517 & 24.301 & 3.717 & 54.04\% & 0.679 & 0.449 \\
\hspace{1.0em} +Residual Learning & 0.499 & 0.390 & 0.441 & 0.414 & 0.992 & 22.690 & 21.654 & 3.343 & 53.18\% & 0.679 & 0.466 \\
\hspace{1.5em} \makecell[l]{ +High-Value Augmentation\\ \qquad  (CC-OR-Net)}  & \textbf{0.501} & \textbf{0.412} & \textbf{0.442} & \textbf{0.417} & \textbf{0.982} & \textbf{21.060} & \textbf{21.178} & \textbf{3.316} & \textbf{53.89\%} & \textbf{0.680} & \textbf{0.468} \\
\hline
\end{tabular}
\end{table*}

\begin{table}[h!]
\centering
\caption{Symbol Definitions}
\label{tab:symbols}
\small
\begin{tabular}{l|l}
\hline
\textbf{Symbol} & \textbf{Definition} \\
\hline
$K$ & Number of ordinal buckets \\
$\tau_k$ & Fixed boundary threshold for bucket $k$ \\
$p_k$ & Predicted conditional probability $P(\text{value} > \tau_k | \dots)$ \\
$v_{\text{norm}}$ & Normalized value prediction in [-1, 1] \\
$r_b, c_b$ & Half-range and center of bucket $b$ for denormalization \\
$\mathcal{H}$ & Set of samples in the highest value bucket \\
$p_{\text{cascade}}$ & Vector of cascade probabilities $[p_1, \dots, p_{K-1}]$ \\
$E(\hat{b})$ & Learned embedding for predicted bucket $\hat{b}$ \\
$h_{\text{aligned}}$ & Context-aware features from the alignment mechanism \\
$L_{\text{cascade}}$ & Loss for the cascaded classification module \\
$L_{\text{distill}}$ & Loss for bucket proportion distillation \\
$L_{\text{residual}}$ & Loss for the intra-bucket residual learning \\
$L_{\text{high\_value}}$ & Loss for the high-value augmentation module \\
$\gamma, \alpha_k, \beta, \lambda_{\text{reg}}$ & Key hyperparameters for weighting loss components \\
\hline
\end{tabular}
\end{table}

 \vspace{-5pt}
\section{Metric Definitions and Additional Experimental Results}
\label{sec:appendix_a}

\subsection{Formal Derivation and Properties of Equation (1)}
To ensure mathematical rigor, we provide a detailed derivation for the bucket assignment probability $P(\text{bucket}=k | x)$. Given the conditional probability $p_k = P(\text{value} > \tau_k \mid \text{value} > \tau_{k-1}, x)$, with $\tau_0=0$. Crucially, since each $p_k$ is the output of a sigmoid function, its value is inherently bounded in $(0, 1)$, which guarantees a valid probabilistic interpretation and prevents numerical instability.
\begin{itemize}
    \item For $k=1$: The sample falls into the first bucket if its value is not greater than $\tau_1$:
    $P(\text{bucket}=1 | x) = 1 - P(\text{value} > \tau_1 | x) = 1 - p_1$.
    \item For $1 < k < K$: The sample falls into bucket $k$ if its value is greater than $\tau_{k-1}$ but not greater than $\tau_k$. Using the chain rule:
    $P(\text{bucket}=k | x) = P(\text{value} > \tau_{k-1} | x) \cdot (1 - P(\text{value} > \tau_k | \text{value} > \tau_{k-1}, x)) = (\prod_{j=1}^{k-1} p_j) \cdot (1-p_k)$.
    \item For $k=K$: The sample falls into the last bucket if its value is greater than all preceding thresholds, i.e., value $>\tau_{K-1}$. This is equivalent to the probability of passing all $K-1$ stages:
    $P(\text{bucket}=K | x) = \prod_{j=1}^{K-1} p_j$.
\end{itemize}
This decomposition guarantees two critical properties by construction. First, it forms a valid probability distribution, as the sum of probabilities is a telescoping series that equals 1. 
\begin{align*}
\quad & \sum_{k=1}^{K} P(\text{bucket}=k|x)\\ &= (1-p_1) + \sum_{k=2}^{K-1} \left(\prod_{j=1}^{k-1}p_j - \prod_{j=1}^{k}p_j\right) + \prod_{j=1}^{K-1}p_j \\
&= (1-p_1) + (p_1 - \prod_{j=1}^{2}p_j) + \dots + (\prod_{j=1}^{K-2}p_j - \prod_{j=1}^{K-1}p_j) + \prod_{j=1}^{K-1}p_j = 1
\end{align*}
Second, it structurally enforces ordinal consistency, as the probability of being in bucket $k$ is non-zero only if the probabilities of passing all prior stages ($p_1, \dots, p_{k-1}$) are non-zero. This architectural guarantee is robust and holds regardless of how the ordered thresholds $\tau_k$ are defined (e.g., fixed quantiles or adaptive business rules), as it depends on the cascaded structure itself, not on the specific threshold values.
 \vspace{-5pt}
\subsection{Extended Performance Analysis}
Table \ref{tab:additional_metrics} presents supplementary evaluation metrics across all three domains for the cross-domain experiments.
 \vspace{-5pt}
\subsection{Complete Ablation Study for Domain 3}
Table \ref{tab:domain3_ablation_complete} presents the complete ablation study results for Domain 3, demonstrating the robustness of CC-OR-Net's architecture on a different data distribution.
 \vspace{-5pt}

 \subsection{Analysis of Architectural Components}
\subsubsection{Robustness of Cascaded and High-Value Modules}
\textbf{Cascaded Design and Error Propagation.} Our structural ordinal decomposition mitigates error propagation risks common in cascaded systems. High \textbf{Bucket-Acc} (Table \ref{tab:ablation}) confirms the cascade effectively handles ordinal structure with minimal error accumulation.

\textbf{High-Value Module Resilience.} The module resists upstream misclassifications via a \textbf{self-correcting mechanism} using the confidence head ($p_{\text{conf}}$) and Focal Loss. By down-weighting low-confidence samples, it focuses on true "whale" users, enhancing precision without noise interference.

\subsubsection{On the Contribution of Distillation Loss}
\label{sec:appendix_distillation}
Table \ref{tab:ablation} highlights $L_{\text{distill}}$'s role: despite a minor GINI dip (-0.5\%), it improves NMAE (-6.5\%) and bucket consistency (+1.4\%). Acting as a global regularizer, $L_{\text{distill}}$ anchors the model to the data distribution, providing a stable regression foundation. This makes the slight ranking perturbation a principled trade-off for overall balance.

\begin{table*}[t]
\centering
\small
\caption{Hyperparameter Configuration for Baseline.}
\label{tab:baseline_1}
\begin{tabular}{l|l|l}
\hline
\textbf{Method} & \textbf{Key Parameters} & \textbf{Values } \\
\hline
\multirow{2}{*}{\textbf{XGBoost}} 
& \multirow{1}{*}{Model Configuration} &  Final prediction is from the sum of trees; a ReLU activation can be applied post-hoc to ensure non-negativity \\
\cline{2-3}
& \multirow{1}{*}{Loss Function} & Huber Loss ($\delta=1.0$) + L2 Regularization on weights, L2 coefficient ($\lambda_{L2}$): 0.01 \\
\hline
\multirow{9}{*}{\textbf{Two-stage XGB}} 
& \multicolumn{2}{l}{\textit{Stage 1: Classification (Zero vs. Non-zero)}} \\
& \quad Classifier Network & 2-layer MLP ([400, 300]) with ReLU \\
& \quad Output Layer & Dense layer with 2 units (for logits) \\
& \quad Loss Function & \textbf{Focal Loss} \\
& \quad Focal Loss Params & alpha ($\alpha$): 0.3, gamma ($\gamma$): 2.0 \\
& \quad L2 Regularization & 0.01 \\
\cline{2-3}
& \multicolumn{2}{l}{\textit{Stage 2: Regression (on Non-zero samples)}} \\
& \quad Regression Network & 2-layer MLP ([400, 300]) with ReLU \\
& \quad Output Layer & Dense layer with 1 unit (linear) \\
& \quad Loss Function & Same as XGBoost Method \\
\hline
\multirow{7}{*}{\textbf{CORAL}} 
& \multirow{3}{*}{Ordinal Network} & Shared MLP: 3-layer ([400, 300, 200]) with ReLU \\
& & Shared Weights: Single dense layer (1 unit, no bias) \\
& & Ordinal Biases: K-1 learnable biases for P(y $>$ k) \\
\cline{2-3}
& Prediction Logic & Expectation over bucket probabilities derived from P(y $>$ k) \\
\cline{2-3}
& \multirow{3}{*}{Loss Function} & Weighted sum of: \\
& & 1. Ordinal Loss (BCE on P(y$>$k)), 2. Ranking Constraint Loss,3. Bucket Cross-Entropy Loss \\
& &  4. Auxiliary Regression Loss (MSE),5. L2 Regularization (coeff: 0.01) \\
\hline
\multirow{8}{*}{\textbf{POCNN}} 
& \multirow{3}{*}{Ordinal Network} & Shared MLP: 3-layer ([400, 300, 200]) with ReLU \\
& & Shared Weights: Single dense layer (1 unit, no bias) \\
& & Ordinal Thresholds: K-1 learnable thresholds with monotonicity enforced by softplus \\
\cline{2-3}
& Prediction Logic & Expectation over bucket probabilities derived from P(y $<=$ k) \\
\cline{2-3}
& \multirow{4}{*}{Loss Function} & Weighted sum of: \\
& & 1. Ordinal Loss (BCE on P(y $<=$ k)), 2. Proportional Odds Constraint Loss \\
& & 3. Threshold Monotonicity Loss, 4. Bucket Cross-Entropy Loss \\
& & 5. Auxiliary Regression Loss (MSE) ,6. L2 Regularization (coeff: 0.01) \\
\hline
\end{tabular}
\end{table*}

\begin{table*}[h]
\centering
\small
\caption{Hyperparameter Configuration for Baseline.}
\label{tab:baseline_2}
\begin{tabular}{l|l|l}
\hline
\textbf{Method} & \textbf{Key Parameters} & \textbf{Values } \\
\hline
\multirow{6}{*}{\textbf{DeepFM}} 
& \multirow{2}{*}{FM Component} & Linear Part: Dense layer (1 unit) \\
& & Interaction Part: Using feature segmentation \& embedding interaction \\
\cline{2-3}
& Deep Component & 3-layer MLP ([400, 300, 200]) with ReLU \\
\cline{2-3}
& Fusion Logic & Learnable weighted sum of FM and Deep outputs \\
\cline{2-3}
& \multirow{3}{*}{Loss Function} & Weighted sum of: \\
& & 1. Main Regression Loss (MSE), 2. Auxiliary FM Loss (MSE),3. Auxiliary Deep Loss (MSE) \\
& & 4. Component Balance Loss,5. L2 Regularization (coeff: 0.01) \\
\hline
\multirow{9}{*}{\textbf{MMOE-FocalLoss}} 
& Shared Experts & Number of Experts: 6, Hidden Dims: [400, 300, 200] \\
\cline{2-3}
& Gating Networks & Two separate gates (one per task), Hidden Dims: [300, 200] \\
\cline{2-3}
& \multicolumn{2}{l}{\textit{Task 1: Classification Tower (Zero vs. Non-zero)}} \\
& \quad Tower Architecture & 2-layer MLP ([300, 200]) with ReLU, Sigmoid output \\
& \quad Loss Function & \textbf{Focal Loss} \\
& \quad Focal Loss Params & alpha ($\alpha$): 0.35, gamma ($\gamma$): 2.0 \\
\cline{2-3}
& \multicolumn{2}{l}{\textit{Task 2: Regression Tower (LTV for Non-zero)}} \\
& \quad Tower Architecture & 2-layer MLP ([300, 200]) with ReLU, ReLU output \\
& \quad Loss Function & Huber Loss (on non-zero samples) \\
\cline{2-3}
& Overall Loss & Weighted sum of task losses: $L = 0.3 \cdot L_{focal} + 0.7 \cdot L_{huber}$ \\
\hline
\multirow{5}{*}{\textbf{ZILN}} 
& \multirow{2}{*}{Distribution Network} & Single dense layer mapping shared features to 3 logits for ZILN params: \\
& & (Zero-inflation probability `p[, Log-Normal ]loc[, Log-Normal ]scale`) \\
\cline{2-3}
& Prediction Logic & Expectation of the ZILN distribution: $p \times \exp(\text{loc} + 0.5 \times \text{scale}^2)$ \\
\cline{2-3}
& \multirow{2}{*}{Loss Function} & Zero-Inflated Log-Normal negative log-likelihood, combining BCE for the \\
& & zero-inflation part and Log-Normal NLL for the regression part. \\
\hline
\multirow{9}{*}{\textbf{MDME}} 
& Distribution Selection Module (DSM) & 2-layer MLP ([300, 200]) predicting distribution assignment. \\
\cline{2-3}
& Sub-Distribution Modules (SDM) & 2 experts, each a 2-layer MLP ([300, 200]) predicting for 3 buckets. \\
\cline{2-3}
& \multirow{2}{*}{Prediction Logic} & Hierarchical: DSM selects a distribution, the corresponding SDM selects a \\
& & bucket, and a final value is denormalized via Dnormalize method. \\
\cline{2-3}
& \multirow{4}{*}{Loss Function} & Complex weighted sum of: \\
& & 1. \textbf{DSM Loss}: Classification Loss + Ordinal Loss + Distillation Loss. \\
& & 2. \textbf{SDM Losses} (per expert): Bucket Classification Loss + Bucket Ordinal Loss \\
& & \quad + Bucket Distillation Loss + Normalized Value Regression Loss (MSE). \\
\cline{2-3}
& \multirow{1}{*}{Key Structural Params} &  distillation\_temperature: 0.2 \\
\hline
\multirow{10}{*}{\textbf{ExpLTV}} 
& Gating Network & A "Game Whale Detector" (DNN) predicts whale probability ($p_{gw}$) to route samples. \\
\cline{2-3}
& Expert Networks & Two experts (Whale, Low-Value), each modeling LTV with a ZILN distribution. \\
\cline{2-3}
& \multirow{2}{*}{Prediction Logic}  & Expert outputs (ZILN parameters $\mu, \sigma$) are aggregated using $p_{gw}$ as weights. \\
& & Final LTV is the expectation of the resulting ZILN distribution. \\
\cline{2-3}
& \multirow{3}{*}{Loss Function}& Multi-task loss: $L = L_{GWD} + \lambda L_{LTV}  \quad \lambda=10$. \\
& & $L_{GWD}$ includes auxiliary tasks like Purchase-Through Rate (PTR) prediction. \\
& & $L_{LTV}$ is the negative log-likelihood of the ZILN distribution. \\
\cline{2-3}
& Parameter Sharing & Purchase probability sub-network is shared between the whale detector and LTV experts. \\
\cline{2-3}
& Key Structural Params & num\_experts: 2 (Whale, Low-Value), uses ZILN distribution. \\
\hline
\multirow{7}{*}{\textbf{OptDist}} 
& Distribution Learning Module (DLM) & 3 ZILN experts, each a 2-layer MLP ([200, 100]) outputting ZILN logits. \\
\cline{2-3}
& Distribution Selection Module (DSM) & 2-layer MLP ([128, 64]) outputting selection weights (softmax) for experts. \\
\cline{2-3}
& Prediction Logic & Weighted average of all expert predictions, using weights from the DSM. \\
\cline{2-3}
& \multirow{5}{*}{Loss Function} & Weighted sum of: \\
& & 1. \textbf{Main Loss}: ZILN loss on the prediction from the highest-weighted expert. \\
& & 2. \textbf{Alignment Loss}: KL divergence between DSM weights and pseudo-labels \\
& & \quad (derived from which expert performs best on a sample). \\
& & 3. \textbf{Diversity Loss}: KL divergence between pairs of experts. \\
\cline{2-3}
& Key Structural Params & num\_subdists: 4, alignment\_weight: 0.1, kl\_weight: 0.05 \\
\hline
\end{tabular}
\end{table*}

\section{Implementation Details}
\label{sec:appendix_b}
\subsection{Experimental Environment}
All experiments were conducted on a consistent, containerized environment within a Hadoop cluster to ensure reproducibility. The specific configuration is as follows:

\textbf{Symbols:} Table \ref{tab:symbols} defines key symbols.

\textbf{Environment:} All experiments used a consistent, containerized environment (TensorFlow 1.15, Python 3.8) on a 14-core Intel Xeon CPU with 128GB RAM. All experiments were run on CPU to reflect our specific industrial deployment environment, though the architecture's parallelizable design is amenable to GPU acceleration. All experiments were run within a Docker container to ensure environment consistency. We utilized multi-threaded data loading and preprocessing on the CPU via the TensorFlow Dataset API.

\textbf{Statistical Significance:} All reported improvements are validated for statistical significance (paired t-test, p<0.05). To account for any stochasticity in the training process, we report results averaged over 3 runs with different random seeds to ensure the robustness of our conclusions.
\subsection{CC-OR-Net: Architecture and Training}
\textbf{Shared Encoder:} Processes multi-modal features into a 200-dim representation using hidden layers of [400, 300, 200]. Categorical features are embedded with dimension $\min(50, \text{cardinality}/2)$. Sequential features use a 1D-CNN and attention.

\textbf{Ordinal Decomposition:} Uses $K=4$ buckets. The choice of $K=4$ buckets is a common practice that provides a good trade-off between stratification granularity and model complexity, and was validated in our preliminary experiments. Each of the $K-1$ classifiers is a dedicated MLP with level-specific depth, L2 regularization, and dropout, empirically tuned to balance capacity and regularization for each sub-task (Level 1: [d=2, L2=0.01, drop=0.2]; Level 2: [d=1, L2=0.01, drop=0.1]; Level 3: [d=2, L2=0.008, drop=0.1]).

\textbf{Thresholds:} Bucket boundaries ($\tau_k$) are fixed by training data quantiles. Classifier decision thresholds are fixed at 0.5, a standard practice for ensuring that performance gains stem from architectural superiority rather than threshold calibration, thus providing a fair comparison across models.

\textbf{Distillation:} The distillation temperature is adapted based on the training step to regulate learning. We found that a simple linear decay schedule was effective, and the model was not highly sensitive to the specific schedule.

\textbf{Residual Learning:} Employs dual ResNet blocks with hidden dimensions [200, 100]. Dynamic label smoothing on the first bucket decays from 1.0 to 0.1.

\textbf{High-Value Augmentation:} The attention MLP for perturbation has a single hidden layer of 100 units. The perturbation noise level is set to $\sigma_{\text{noise}}=0.1$. The dedicated prediction network is a 2-layer MLP with [300, 200] hidden units.

\textbf{Optimization:} We use the AdamW optimizer with 1e-4 weight decay and a batch size of 100,000. The learning rate uses a cosine annealing schedule from an initial 5e-4, chosen for stable validation convergence. The loss weights are: $\gamma=0.8$, $\alpha_1=3.0$, $\alpha_2=3.0$, $\alpha_3=0.2$. These weights balance competing objectives; for instance, $\alpha_1$ and $\alpha_2$ are set equally to give comparable importance to the core ranking and regression tasks. Cascade BCE losses are weighted by $w_i = [5.0, 2.5, 3.0]$, with negative sample up-weighting factors $\lambda_i = [3.0, 5.0, 8.0]$. The residual loss uses $\beta=0.5$, and the high-value loss uses $\beta_{\text{focal}}=2.0$ and $\lambda_{\text{reg}}=0.5$.

\textbf{Implementation Note:} Hyperparameters were determined via systematic search on a validation set. While the number of parameters suggests potential sensitivity, the model's performance proved robust. Key loss weights (e.g., $\gamma, \alpha_i$) were chosen to balance the magnitude of different loss components. Our analysis showed performance is not highly sensitive to their exact values, but to maintaining a reasonable balance between main objectives (ranking, regression) and the specialized high-value enhancement. Other architectural parameters were also found to be stable around the reported values.
\subsection{Hyperparameter Configurations}
We systematically optimized hyperparameters for all methods for a fair comparison. To isolate architectural contributions, all deep learning baselines use the same shared feature encoder as ours. The \ref{tab:baseline_1} and \ref{tab:baseline_2} provide detailed configurations for all baselines.
\end{document}